\documentclass[6pt,journal]{IEEEtran}
\usepackage[latin1]{inputenc}
\usepackage[normalem]{ulem}
\usepackage{indentfirst}
\usepackage{color}
\usepackage{eucal}
\usepackage{multirow}
\usepackage{soul} 

\usepackage{float}
\usepackage{amssymb,amsmath}
\usepackage{hyperref}   
\usepackage{amsthm}   
\usepackage{flushend}
\usepackage{graphicx}
\usepackage{stfloats}
\usepackage{comment}

\usepackage{cite}  

\usepackage[framemethod=tikz]{mdframed}
\usepackage{lipsum}
\usepackage{tikz}
\usepackage{rotating}
\usetikzlibrary{shapes.geometric, arrows}
\usepackage{adjustbox}
\usepackage{verbatim}

\newtheoremstyle{normalnoit}{}{}{}{}{\bf }{}{ }{}
\theoremstyle{normalnoit}

\newcommand{\be}{\begin{equation}}
\newcommand{\ee}{\end{equation}}
\newtheorem{theorem}{Theorem}[section]

\newtheorem{proposition}[theorem]{Proposition}
\newtheorem{remark}[theorem]{Remark}

\ifCLASSINFOpdf
\ifCLASSINFOpdf
\else
\fi
\newcommand{\orcidicon}{\includegraphics[width=0.32cm]{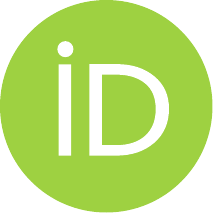}}
\foreach \x in {A, ..., Z}{%
\expandafter\xdef\csname orcid\x\endcsname{\noexpand\href{https://orcid.org/\csname orcidauthor\x\endcsname}{\noexpand\orcidicon}}
}

\begin{document}

\title{A Low-complexity Structured Neural Network Approach to Intelligently Realize Wideband Multi-beam Beamformers
}
%
\author{Hansaka Aluvihare\orcidA{}, Sivakumar Sivasankar,
	Xianqi Li\orcidC{}, Arjuna Madanayake\orcidD{}, and Sirani M. Perera\orcidE{}.
\thanks{H. Aluvihare is with the Department
of Mathematics, Embry-Riddle Aeronautical University, Daytona Beach,
FL, 32703 USA e-mail:aluvihah@my.erau.edu}
\thanks{S. Sivasankar is with the Department of Electrical and Computer Engineering, Florida International University, Miami, FL, 33174 USA e-mail:ssiva011@fiu.edu}
\thanks{X. Li is with the Department
of Mathematics \& Systems Engineering, Florida Institute of Technology, Melbourne, FL 32901, USA e-mail: xli@fit.edu (see https://www.fit.edu/faculty-profiles/l/li-xianqi/).}
\thanks{A. Madanayake is with the Department
of Electrical \& Computer Engineering, Florida International University, Miami, FL 33174 USA e-mail: amadanay@fiu.edu (see https://ece.fiu.edu/people/faculty/profiles/madanayake-arjuna/).}
\thanks{S. M. Perera is with the Department
of Mathematics, Embry-Riddle Aeronautical University, Daytona Beach,
FL, 32703 USA e-mail: pereras2@erau.edu (see https://faculty.erau.edu/Sirani.Perera).}

\thanks{This work was supported by the National Science Foundation award numbers 2229473 and 2229471.}
}
\maketitle



\begin{abstract}
True-time-delay (TTD) beamformers can produce wideband, squint-free beams in both analog and digital signal domains, unlike frequency-dependent FFT beams. Our previous work showed that TTD beamformers can be efficiently realized using the elements of delay Vandermonde matrix (DVM), answering the longstanding beam-squint problem. Thus, building on our work on classical algorithms based on DVM, we propose neural network (NN) architecture to realize wideband multi-beam beamformers using structure-imposed weight matrices and submatrices. The structure and sparsity of the weight matrices and submatrices are shown to reduce the space and computational complexities of the NN greatly. The proposed network architecture has $\mathcal{O}( p L M \log M )$ complexity compared to a conventional fully connected $L$-layers network with $\mathcal{O}(M^2L)$ complexity, where $M$ is the number of nodes in each layer of the network, $p$ is the number of submatrices per layer, and $M > >  p$. We will show numerical simulations in the 24 GHz to 32 GHz range to demonstrate the numerical feasibility of realizing wideband multi-beam beamformers using the proposed neural architecture. We also show the complexity reduction of the proposed NN and compare that with fully connected NNs, to show the efficiency of the proposed architecture without sacrificing accuracy. 
The accuracy of the proposed NN architecture was shown using the mean squared error, which is based on an objective function of the weight matrices and beamformed signals of antenna arrays, while also normalizing nodes.
The proposed NN architecture shows a low-complexity NN realizing wideband multi-beam beamformers in real-time for low-complexity intelligent systems.
\end{abstract}


\section{Introduction}\label{sec: Intro}
Beamforming has been widely explored for its diverse applications across fields, such as radar, communication, and imaging. Transmit beamforming overcomes path loss by concentrating energy into a specific direction while receiving beamforming directionally enhances propagating planar waves based on a desired direction of arrival~\cite{ref1}. 
When the signal of interest is wideband, multi-beam beamforming based on the spatial Fast Fourier transform (FFT) suffers from the beam-squint problem \cite{ref2}.
\subsection{Realize TTD-based Beamformers via DVM}
FFT-beams are frequency dependent and thus cause poor beam orientations for wideband signals. Fortunately, the true-time-delay (TDD) beamformers have significantly mitigated the beam-squint problem associated with spatial FFT beams \cite{ref6}. On the other hand, the Vandermonde delay matrix (DVM) elements can be utilized to determine the TTD beams \cite{ref2, ref3, ref4, ref6}. This amounts to incorporating the DVM between antennas and source/sink channels and implementing via frequency-dependent phase shifts at each antenna to achieve TDD beamformers leading to wideband multi-beam architecture. 
Thus, utilizing TDD, at time $t \in \mathbb{R}$ the $N$-beam
beamformer $\tilde{y}$ can be expressed as a product of the input vector $\tilde{x}$ and the DVM $A_N$, s.t., $\tilde{y}=A_N \tilde{x}$ .
 In this context, each row of the DVM matrix symbolizes the progressive wideband phase shift associated with a specific beam. However, computational cost plays a crucial role in computing the matrix-vector multiplication associated with wideband multi-beam beamformers. Each TTD is typically realized in the digital domain using a finite impulse response (FIR) digital filter - sometimes known as a Frost Structure. Thus, in order to reduce the delays of $N$ beams from $\mathcal{O}(N^2)$ to $\mathcal{O}(N \log N)$, we proposed sparse factorization to realize narrowband multi-beam beamformers \cite{ref3}, and 
 wideband multibeam beamformers \cite{ref6}.  
The necessity of retaining intermediate values in memory can result in increased memory demands. Such circumstances may pose a disadvantage in real-time or low-latency applications. 
Hence, a critical requirement emerges for the development of a real-time training and prediction algorithm to effectively realize wideband multi-beam beamforming. Thus, we propose employing shallow and fully connected NNs to realize
wideband multi-beam beamformers while imposing structures for weight matrices to propose a lightweight and low-complexity NN so that we could show numerical simulations for .

\subsection{Neural Networks Approaches for Beamformers}
Several methods have been proposed for the application of both shallow and deep neural networks or multi-layer perceptrons in the context of adaptive beamforming as applied to phased arrays \cite{ref80,ref81,zaharis2016implementation,sallomi2016multi}. In \cite{ref7}, a radial basis function neural network (RBFNN) was employed to approximate the beamformers derived through the application of a minimum mean-squared error (MSE) beamforming criterion while adhering to a specified gain constraint. In \cite{ref8}, a NN was trained to create adaptive transmit and receive narrowband digital beamformers for a fully digital phased array. Many convolutional neural network (CNN) based adaptive algorithms have been proposed, such as  \cite{rs15030712, liao2023robust, 9266516, ref9, 9259040}. In \cite{rs15030712}, an approach known as frequency constraint wideband beamforming prediction network (WBPNet) is introduced without delay structure based on a CNN method to  tackle the limitations associated with insufficient received signal snapshots while reducing computational complexity. This CNN-based method focused on predicting the direction of arrival (DOA) of interference. Then \cite{9259040} introduce a CNN-based neural beamformer to predict the interference from received signals and an LSTM model to predict the samples of desired signals for a low number of receiving snapshots. In \cite{sallam_attiya_2021}, a CNN is trained based on the data obtained from the optimum Wiener solution and results are compared with $8 \times 8$ antenna arrays. Moreover, in \cite{8756639} a scheme is introduced to predict a power allocation vector before determining the beamforming matrix with CNN. This method addresses the challenge of overly complex networks and power minimization problems in the context of wideband beamforming for synthetic aperture radar (SAR). Above methods include training of a CNN model to design the beamformer for specific sizes of antenna arrays. However, as the number of elements in the array increases (which is expected for mmWave communications) there is a lack of research that  evaluates the relative performance of the above methods. The  authors of \cite{8847377} proposed a multilayer neural network model to design a beamformer for 64-element arrays to tackle the challenges in imperfect CSI and hardware challenges by maximizing the spectral efficiency. Besides, \cite{8888573} proposed a CNN-based beamformer to estimate the phase values for beamforming. Furthermore, \cite{9762632} and \cite{che2016recurrent} explored the recurrent neural network-based algorithm to estimate the weights in the antenna array. Authors in \cite{9762632} proposed GRU-based ML algorithms for adaptive beamforming.

\subsection{Structured Weight Matrices in Neural Networks}
As modern NN architectures grow in size and complexity, the demand for computational resources is significantly increasing. Structured weight matrices present a solution to mitigate this increased resource consumption by simplifying computational tasks \cite{sze2017efficient}. These matrices, by leveraging inherent structures, can reduce the computational complexity for propagating information through the network \cite{kissel2023structured}. 
However, selecting the appropriate structure within the diverse array of matrix structures and classes is not a trivial task. To address this challenge, numerous methods \cite{feng2015learning, gong2014compressing, han2015deep, wen2016learning, zhao2017theoretical, kamalakara2022exploring, lingsch2023structured, liao2019circconv} have been developed to minimize the computational costs and memory requirements of neural networks. Those existing efforts generally fall into two categories: reduction techniques focused on fully-connected NN including weight pruning/clustering \cite{gong2014compressing, han2015deep}, which prune and cluster the weights via scalar quantization, product quantization, and residual quantization, to reduce the NN model size, and reduction strategies aimed at convolutional layers, such as low-rank approximation \cite{sainath2013low, jaderberg2014speeding, kamalakara2022exploring} and sparsity regularization \cite{feng2015learning, wen2016learning}. These approaches are critical for enhancing the efficiency of neural networks, making them more practical for a variety of applications.

\subsection{Objective of the Paper}
Our goal is to introduce a structure-imposed NN (StNN) to realize multi-beam beamformers while dynamically updating the StNN with a low-complexity and lightweight NN. 
We have shown that the sparse factorization of the DVM is an efficient strategy to reduce the complexity in computing the DVM-vector product from $\mathcal{O}(N^2)$ to $\mathcal{O}(N \log N)$.  
Nevertheless, it is crucial to dynamically update delays and sums based on the DVM-vector product so that we can intelligently realize multi-beam beamformers.     
Fortunately, we could regularize the weight matrices using NNs while adopting the sparse factorization of the DVM in \cite{ref6}, and train, update, and learn TTD beamformers while imposing the structure of the DVM followed by the sparse factors. Hence,  we propose a hybrid of classical and ML algorithms to dynamically realize multi-beam beamforming, in contrast to weight-pruning techniques that result in irregular pruned networks \cite{anwar2017structured}. Since the DVM can be fully determined using the parameters $\mathcal{O}(N)$ and the DVM vector product can be computed with $\mathcal{O}(N \log N)$ complexity, the factorization of the DVM imposed as weight matrices within the NNs could greatly reduce computational complexity. The proposed StNN architecture leads to 
\begin{enumerate}

\item intelligently realizes wideband multibeam beamformers while reducing TTD blocks, 

\item ensures a robust structure for the trained network while reducing computational complexities incurred by complex indexing processes,

\item reduce computational complexities due to the usage of structured and sparse weight matrices, i.e., 70\% complexity reduction compared to our previous paper \cite{ASPSKAL24}, and
 
\item obtain a lightweight NN while intelligently realizing wideband multibeam beamformers.
\end{enumerate}
We note here that the DVM is a low displacement rank (LDR) matrix, and LDR-based neural networks have gained attention due to their potential to reduce complexity when the structure is imposed for the neural network \cite{zhao2017theoretical, kamalakara2022exploring,lingsch2023structured}.
%
%
%
Thus, the utilization of the DVM structure followed by factorization of the low-rank DVM in \cite{ref6}, without the need for retraining (due to utilization of frequencies, i.e., 24, 27, \& 28 GHz) 
lead to propose a low-complexity StNN that can be utilized to intelligently realize wideband multi-beam beamformers. 
%

\subsection{Structure of the Paper}
The remainder of the paper is organized as follows. Section~\ref{sec: Methodology} introduces the theory of the structure-imposed neural network model to realize wideband multi-beam beamformers. Section~\ref{sec: Complexity Analysis} shows the arithmetic complexity of the StNN showing the reduction of the complexity. 
and Section~\ref{sec: Simulation} shows the numerical simulations showing the efficiency and accuracy of the proposed StNN as opposed to the fully connected neural network in realizing wideband multibeam beamformers in 24 GHz to 32 GHZ range. Finally, the Section~\ref{sec: Conclusion} concludes the paper.

\section{Methodology}\label{sec: Methodology}
\begin{figure*}[htp]
\centering
\centerline{\includegraphics[width=175mm]{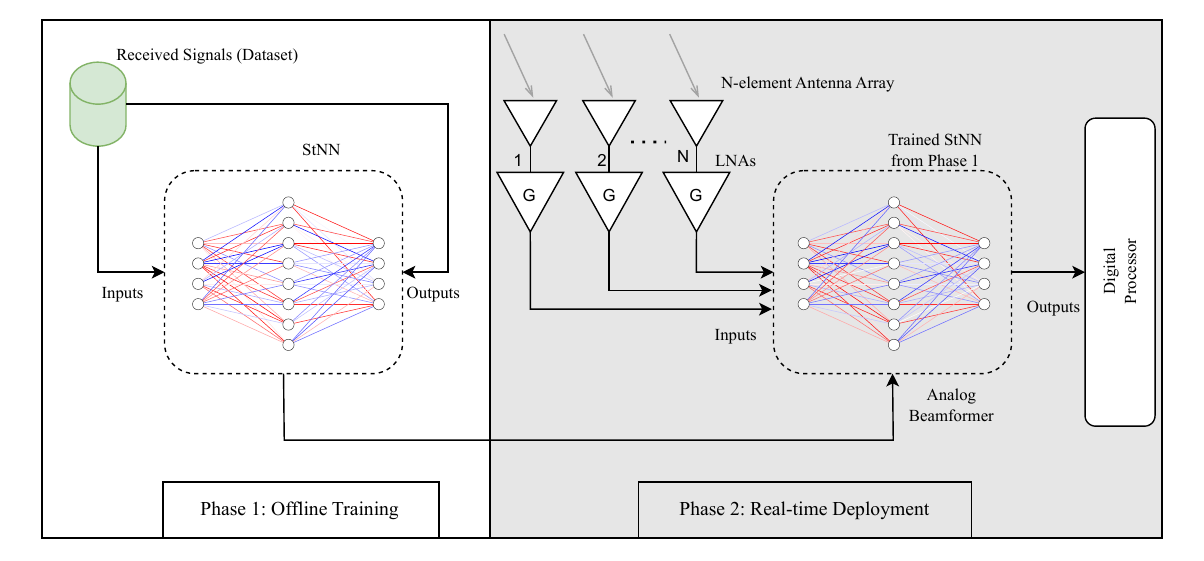}}
\caption{ML-based architecture of multi-beam beamforming: In the offline training, we train the neural network to align the input data by the weight matrix to the desired output data. In real-time deployment: RF signals from the antennas and low noise amplifiers (LNAs) are beamformed utilizing the structure imposed neural network, i.e., StNN. Once the multibeams are formed they will be sent to the digital processor.}
\label{fig:high level design}
\end{figure*}

We first present the theoretical framework for the Structured Neural Network (StNN) before numerically realizing wideband multi-beam beamformers. The design of StNN utilizes customized weight matrices to effectively incorporate the structure of the DVM, followed by a sparse factorization in \cite{ref6}. This approach enables efficient computation of DVM-vector multiplications, enabling the realization of wideband multi-beam beamformers, and ensuring model stability. Unlike conventional feed-forward neural networks(FFNN), StNN significantly reduces the computational complexity. It optimizes space and storage requirements, handling large-scale systems involving high values of $N$, making it a more scalable, low-complexity data-driven alternative to conventional multi-beam beamformers.

The high-level design of the proposed framework is shown in Figure \ref{fig:high level design}, illustrating a two-phase process comprising offline training and real-time deployment. During real-time operation, the StNN processes input data from each antenna element to estimate the multibeam beamformer output. Prior to deployment, the StNN model undergoes offline training on a pre-collected dataset of received RF signals. After training, the StNN predicts $N$ beamformer output signals based on the true time-delay Vandermonde beamformer, which the digital processor then uses for further processing.

Previous research \cite{zhao2017theoretical} introduced the concept of leveraging low displacement rank (LDR) structured matrices in NNs to reduce both storage and computational overhead. This was achieved through factorization via matrix displacement equations. In contrast, our approach employs a StNN leveraging the DVM factorization instead of displacement equations. This strategic choice effectively minimizes the number of trainable weights and inference-time floating-point operations (FLOPs), leading to a more efficient and computationally lightweight neural network architecture for multi-beam beamforming applications.

\subsection{DVM Factorization in \cite{ref6}}




The DVM is defined using the node set \({A}_N=\{\alpha^{kl}\}_{k=1, l=0}^{N, N-1}\), where \(\alpha = e^{-j\omega\tau} \in \mathbb{C}\). Here, \(N = 2^r\) for \(r \geq 1\), \(\omega\) represents the temporal frequency, and \(\tau\) denotes the time delay. In \cite{ref6}, we introduced a scaled version of the DVM, denoted as \(\widetilde{A}_N\), which facilitates factorization into sparse matrices. This factorization enables efficient computation of the DVM-vector product using an optimized algorithm with a computational complexity of \(O(N\log(N))\), as expressed in the following equation:

\begin{equation}
\widetilde{A}_N = \hat{D}_N[J_{M\times N}]^T F_M^* \Breve{D}_M F_M J_{M\times N} \hat{D}_N.
\label{eq: DVM fac equation}
\end{equation}

Here, \(M = 2N\), \(\hat{D}_N = \text{diag}[\alpha^{\frac{k^2}{2}}]_{k=0}^{N-1}\) is a diagonal scaling matrix, $\breve{D}_{M}={\rm diag}\left[\tilde{\bf F}_{M}{\bf c} \right]$ where a circulant matrix $C_{M}$ defined by the first column ${\bf c}$ s.t. $
{\bf c} =\left[1, \alpha^{-\frac{1}{2}},  \cdots, \alpha^{-\frac{(N-1)^2}{2}}, 1,  \alpha^{-\frac{(N-1)^2}{2}}, \alpha^{-\frac{(N-2)^2}{2}}, \cdots, \alpha^{-\frac{1}{2}} \right]^T
$, \(J_{M\times N} = \begin{bmatrix} I_N \\ 0_N \end{bmatrix}\) is a zero-padded identity matrix, \(I_N\) denotes the identity matrix, while \(0_N\) represents the zero matrix, The Discrete Fourier Transform (DFT) matrix is given by \(F_N = \frac{1}{\sqrt{N}}[\omega_N^{kl}]_{k,l=0}^{N-1}\), where the nodes are defined as \(\omega_N = e^{-\frac{2\pi j}{N}}\) with \(j^2=-1\) and \(F_M^{*}\) represents the conjugate transpose of the DFT matrix \(F_M\).

This structured factorization not only enhances computational efficiency but also reduces storage and processing complexity, making it a viable approach for large-scale implementations.  

\subsection{DVM Structure-imposed Neural Networks (StNN)}
{To efficiently compute the multi-beam beamformer output, we use the DVM structure and the factorization from (\ref{eq: DVM fac equation}) to impose structure for the weight matrices of the StNN. The proposed StNN follows an $L$-layer feedforward architecture, consisting of an input layer, output layer, and $l$ hidden layers, where $l = L - 2$.} Notably, This framework is adaptable, allowing for the addition of more hidden layers and units to accommodate the accuracy of the predictions.

\begin{figure}[htp]
\centering
\centerline{\includegraphics[width=10cm]{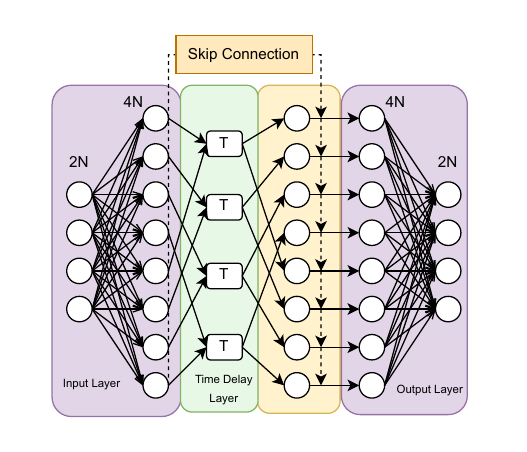}}
\caption{StNN architecture for predicting the output of the TTD beamformers. $2N$ neurons in the input layer (separating real-vales and imaginary parts of the received vector $\tilde{x}$ giving $2N$ neurons in the input vector $\underbar{x} \in \mathbb{R}^{2N}$), $4pN$ neurons in the hidden layer, where  $p, N \in \mathbb{Z}^+$, $4pN$ neurons in each hidden layers and $2N$ neurons in the output vector $\underbar{y} \in \mathbb{R}^{2N}$ resulting the beamformed vector $\tilde{y} \in \mathbb{C}^N$.}
\label{fig:NN}
\end{figure}

The neural network architecture of the proposed StNN model is illustrated in Fig. \ref{fig:NN}. 
Given that the received vector from $N$ elements antenna array consists of complex-valued signals $\tilde{x} \in \mathbb{C^N}$ representing the received RF signals, we separate the real and imaginary components of each signal to pass that to the StNN. This transformation ensures that only real-valued inputs $\underbar{x} \in \mathbb{R}^{2N}$ are processed within the StNN. Each complex number $a+ib$ is mapped to a real-valued pair $(a,b)\in\mathbb{R}^2$. Consequently, for $N$ complex-valued inputs, a corresponding real-valued input vector of size $2N$ is generated for the StNN.
%

\vspace{0.25in}

\subsubsection{Forward Propagation of the StNN}
First, we obtain the forward propagation equations for the StNN. Here, we consider the input layer consisting $2N$ neurons and a fully connected hidden layer with $4pN$ neurons where with a weight matrix $W \in \mathbb{R}^{4pN \times 2N}$, 
where $p$ is the number of submatrices. 
When a $2N$ sized input vector $\underline{x}$ is given to the StNN, the general forward propagation equation for the first hidden layer through a fully connected layer can be expressed as
\begin{equation}
    \underline{y}^{(1)} = \sigma(W^{(1)} \underline{x} + \underline{\theta}^{(1)}),
    \label{eq:generalFP}
\end{equation}
where $\underline{y} \in \mathbb{R}^{4pN}$ is the output of the first hidden layer, $W^{(1)} \in \mathbb{R}^{4pN \times 2N}$ is the weight matrix between the input layer and the first hidden layer 
(defined by the sparse matrices aligned with the factorization equation \ref{eq: DVM fac equation} (as described next), $\theta$ is the bias vector, and $\sigma (.)$ denotes the activation function of the current layer.

In general, the forward propagation equation for the output vector $\underline{y}^{(l+1)}$ for any given $(l+1)$-th hidden layer can be expressed as,
\begin{equation*}
    \underline{y}^{(l+1)} = \sigma(W^{(l+1)} \underline{y}^{(l)} + \underline{\theta}^{(l+1)}),
\end{equation*}
where, $\underline{y}^{(l)}$ is the output of the previous hidden layer. 
Next, we redesign the weight matrices between the layers of the StNN. More precisely, weight matrices, $W^{(1)}$ (i.e. weight matrix between the input layer and the first hidden layer) and $W^{(4)}$(i.e. weight matrix between the last hidden layer and the output layer) in StNN shown in the Fig. \ref{fig:NN}, is decomposed into $p$ smaller sub weight matrices. For instance, the weight matrix between the input layer and the first hidden layer (i.e. $W^{(1)}$), is structured as shown in Equation (\ref{eq: W1_p}). Similarly, the weight matrix between the last hidden layer and the output layer (i.e. $W^{(4)}$) follows the structure presented in Equation (\ref{eq: W4_p}). 

\begin{align}
    W^{(1)} = 
    \begin{bmatrix}
         w^{(1)}_1&
          w^{(1)}_2&
        \cdots & 
        w^{(1)}_p &
    \end{bmatrix}^T
    \label{eq: W1_p}
\end{align}

\begin{align}
    W^{(4)} &= [w^{(4)}_1, w^{(4)}_2, w^{(4)}_3, ..., w^{(4)}_p]
    \label{eq: W4_p}
\end{align}

In the StNN architecture, the weight matrix between the first hidden layer and the second hidden layer, i.e., $W^{(2)}$ is not fully connected and it acts as a physical delay to the signals, meaning it does not contain any trainable weights. Instead, it consists solely of $2pN$ time delay elements. The primary purpose of this layer is to introduce delay to the signal transformed by the first layer. Each delay element applies a fixed delay to the output signal from the first hidden layer. With the introduction of a physical delay layer inside the StNN, StNN evolves to better fit for wideband multi-beam beamformers where time delays are crucial in producing the beamformer output. An important consideration is that the output of the first hidden layer, (i.e. $\underline{y}^{(1)} \in \mathbb{R}^{4pN}$) is a real-valued vector. Therefore, Before applying the delay, this real-valued vector is converted into a complex-valued signal ($\tilde{y}^{(1)}$). This operation can be expressed as follows.
\begin{align*}
    \tilde{\underline{y}}^{(1)} = \underline{y}^{(1)}_{1:2pN} + j\underline{y}^{(1)}_{2pN:4pN}
\end{align*}
Where $\underline{y}^{(1)}_{1:2pN}$ represents the first half of $\underline{y}^{(1)}$ (real part) and $\underline{y}^{(1)}_{2pN:4pN}$ represents the second half of $\underline{y}^{(1)}$, $j$ is the imaginary unit.
The delay is then applied to the reconstructed complex signal $\tilde{\underline{y}}^{(1)} \in \mathbb{C}^{2pN}$ producing $\tilde{\underline{y}}^{(2)} \in \mathbb{C}^{2pN}$,
\begin{align*}
    \tilde{\underline{y}}^{(2)} = W^{(2)}. \tilde{\underline{y}}^{(1)},
\end{align*}
where $W^{(2)} = {\rm diag} \left[\alpha^{k}\right]_{k=0}^{2pN}$.
After $\tilde{\underline{y}}^{(2)}$ is converted back into a real-valued signal $\underline{\underline{y}}^{(2)}$ by separating the real and imaginary components. 
\begin{align*}
    &\underline{y}^{(2)}_{1:2pN} := \tilde{\underline{y}}^{(2)}_{Re}, &\underline{y}^{(2)}_{2pN:4pN} := \tilde{\underline{y}}^{(2)}_{Im},     
\end{align*}
where, $\tilde{\underline{y}}^{(2)}_{Re}$ represents the real components of $\tilde{\underline{y}}^{(2)}$ and $\tilde{\underline{y}}^{(2)}_{Im}$ represents the imaginary components of the $\tilde{\underline{y}}^{(2)}$. 
The resulting output of the second hidden layer (i.e. $\underline{y}^{(2)} \in \mathbb{R}^{4pN}$) is thus a real-valued vector. In summary, the second hidden layer is a non-trainable layer that does not contain any weights but applies a time delay to the complex signal.

In the third hidden layer, we apply the weighted skip connection to $y^{(1)}$, which is then added to $y^{(2)}$ to produce the $y^{(3)}$. The forward propagation equation for this process is provided below.
\begin{align}
    \underline{y}^{(3)} = \underline{y}^{(2)} + W^{(3)}.\underline{y}^{(1)}
\end{align}
Where, $W^{(3)} = {\rm diag} \left[w_k\right]_{k=0}^{4pN} \in \mathbb{R}^{4pN \times 4pN}$ is a diagonal matrix in which the weights along the diagonal are trained, while all the other weights that are not on the main diagonal remain zero. 

\subsubsection{Structure Imposed Sparse Weight Matrices}
The StNN features trainable weights in $W^{(1)}$, $W^{(3)}$, and $W^{(4)}$, while the weights in $W^{(2)}$ remain frozen. With these structured weight matrices, we achieve a substantial reduction in TTDs, decreasing from $N^2$ in traditional TTD beamformers to $2pN$ in the StNN-based beamformer. This reduction becomes even more significant as $N$ increases, which we will demonstrate in the simulation and results section \ref{sec: Simulation}. However, despite the reduced number of TTDs, the StNN still exhibits computational complexity of $\mathcal{O}(N^2)$ in generating the beamformer output (i.e. This complexity arises due to fully connected weight matrix-vector multiplications involving $W^{(1)}\text{ and } W^{(4)}$ with the input vectors in the corresponding layer). To mitigate this and achieve a reduced complexity, we impose matrix factorization on the fully connected weight matrices, as expressed in Equation (\ref{eq: DVM fac equation}). For each submatrix $i$ (i.e $w^{(1)}_{i}$) in $W^{(1)}$, we employ a split factorization based on Equation (\ref{eq: DVM fac equation}), where the factorization is defined as,
\begin{align}
    w^{(1)}_{i} = [\Breve{D}_i]_{2M} [F_i]_{2M} J_{2M\times 2N} [\hat{D}_i]_{2N}
\end{align}
for $p$ such matrices. This $p$ submatrix approach is necessary because, when transitioning from the input layer to the first hidden layer, the number of nodes must increase to effectively capture patterns among the input features. The increased number of parameters introduced during the DVM factorization facilitates the expansion of input features into a higher-dimensional space within the first hidden layer. A similar factorization strategy is applied to the final layer, utilizing the remaining split DVM factorization from Equation (\ref{eq: DVM fac equation}). Specifically, we employ $p$ matrices of
\begin{align}
    w^{(4)}_{i} = [\hat{D}_i]_{2N}[J_{2M\times 2N}]^T [F^*_i]_{2M}
\end{align}

In summary, we implement the DVM factorization for each $p$ submatrix within the weight matrices, where each submatrix is a product of sparse matrices. The structured $p$ submatrices appear between the input layer and the first hidden layer having each submatrix with a size of $2M \times M$. 
Furthermore, there are $p$ submatrices, and each submatrix with a size of $M \times 2M$ appears between the last hidden layer and the output layer.
Moreover, the training process of submatrices involves learning parameters that are only located along the diagonal in matrices $\hat{D}_i$ and $\Breve{D}_i$ while keeping other values fixed at zero without updates during backpropagation. Additionally, matrix $J$ within the StNN remains frozen, exempt from training adjustments during backpropagation. 


\subsubsection{Recursive Algorithm for Weight Matrices}
In this section, we incorporate the recursive strategy presented in \cite{ref6} to reduce the number of additions and multiplications. The main objective of this approach is to reduce the additions and multiplications so that the total number of adders and multipliers in an AI-based circuit can be reduced. 
For example when $M = 2N$, the matrix $F_i$ { appears within the submatrices } can be factored as follows \cite{ref6}:
\begin{align}
    [\widetilde{F}_i]_{2N} &= P_{2N}
    \begin{bmatrix}
        [\widetilde{F}_{i}]_N  & 0_N\\
        \\ 
        0_N & [\widetilde{F}_{i}]_N
    \end{bmatrix}
    \label{eq:recursive_F_1}
   [H_i]_{2N} \\
   [{H}_i]_{2N} &=
    \begin{bmatrix}
        I_N & I_N \\
        \\
        [\widetilde{D}_{i}]_N  &  [-\widetilde{D}_{i}]_N 
    \end{bmatrix}
    \label{eq:recursive_F_2}
\end{align}
where, $\widetilde{D}_{i}$ is a diagonal matrix with values along the diagonal, $O_N$ is a zero matrix, $I_N$ is a identity matrix and $P_{2N}$ is a $2N \times 2N$ sized even-odd permutation matrix \cite{ref6}.


Thus, utilizing the equations (\ref{eq:recursive_F_1}) and (\ref{eq:recursive_F_2}), we can recursively factorize \([\widetilde{F}_{i}]_{2N}\) matrices. 
Through this recursion, the matrix factorization can be performed up to 
\([\widetilde{F}_{i}]_2\), resulting in \(log(M)\) factorization steps. The determination of factorization steps is based on the performance. Opting for higher steps significantly reduces the weight of the network. However, it may also increase the error of the predicted output due to the reduced number of weights in the NN. 
Hence, there exists a trade-off when the number of recursive factorization steps is a hyperparameter in the StNN model that needs to be tuned based on the results.
In the training process, $P_{2N}, O_N$, and $I_N$ are {fixed} matrices with ones and zeros, and those matrices are not updated through backpropagation. During the recursive factorization, we are only updating the matrices $\widetilde{D}_{i}$  and  \([\widetilde{F}_{i}]_2\),  at each recursive step. Here, $\widetilde{D}_{i}$ diagonal matrix at every step is different and independent of each other, and we allowed the neural network to learn weights. The \([\widetilde{F}_{i}]_2\) matrix is trained as a full matrix to yield \([\widetilde{F}_{i}]_{2N}\) while training \([\widetilde{F}_{i}]_2\) during the backpropagation. As a summary, during backpropagation of the recursive factorization, we only need to train a set of diagonal and \([\widetilde{F}_{i}]_2\) matrices. When we update weights through gradient descent, we only update weights that are along the diagonal elements in each diagonal matrix, while remaining the rest of values as zero.

\begin{remark}
We note here that our work on a structured-based NN architecture could also be found, i.e., a classical algorithm utilized to design layers of NNs, to realize states of dynamical systems in \cite{HLSL25}.     
\end{remark}

\subsubsection{Backpropagation of the StNN}
The backpropagation process in the StNN follows the standard gradient-based optimization framework (PyTorch's automatic differentiation engine - Autograd) to compute gradients efficiently. The proposed StNN architecture is implemented in Python using the PyTorch library, where the gradients of all trainable parameters are automatically computed using the above framework.

During training for diagonal matrices, only the weights along the diagonals are updated, while all off-diagonal elements remain zero and are frozen throughout training. This results in highly sparse weight matrices, significantly reducing the number of trainable parameters while preserving the model's ability to capture essential transformations.
We use the Mean Squared Error (MSE) as the loss function (\ref{eq: objective function}) to update weights via
\begin{align}
   O(W^{(1)}, \cdots, W^{(4)}) = \frac{1}{NM_b} \sum_{s=1}^{M_b} \sum_{k=1}^{N} (y_{s}^{(k)} - \hat{y}_{s}^{(k)})^2,
   \label{eq: objective function}
\end{align}
where $W^{(1)}, \cdots, W^{(4)}$ are defined via (\ref{eq: W1_p}) and (\ref{eq: W4_p} respectively,  $M_b$ is the mini-batch size, $y_s^{(k)}$ and $\hat{y}_s^{(k)}$ denote the actual and predicted values at $k^{th}$ antenna index for the $s^{th}$ data sample, respectively.

\section{Arithmetic Complexity Analysis}\label{sec: Complexity Analysis}
In this section, we present an analysis of the arithmetic complexity of the StNN
having an arbitrary input vector $\underline{x} \in \mathbb{R}^{2N}$, where $M:=2N$, which is constructed by extracting real and imaginary parts of the vector $\tilde{\underline{x}}$). 
In this calculation, we assume that the number of additions $(\# a)$ and multiplications $(\# m)$ required to compute $[F_i]_N$ by an $N$ dimensional vector as $Nr$ and $\frac{1}{2}Nr + \frac{3}{2}N$ \cite{ref2}, respectively, where, $N = 2^r(r \geq 1)$.  
\begin{proposition}
 Let the StNN be constructed using  $L$ layers, i.e., the input layer with $M$ nodes,  $L-2$ hidden layers with $2pM$ nodes consisting $p$ submatrices per hidden layer, and an output layer with $M$ nodes. Then, the number of additions $(\# a)$ and multiplications $(\# m)$ of the StNN having the input vector $\underline{x} \in \mathbb{R}^M$, where, $N = 2^r(r \geq 1)$ and $M=2N$ 
 is given via    
 \begin{align}
    \#a(StNN) = &p(L-1)Mr + 4p(L-1)M -\frac{(L-1)}{4}M \nonumber \\
    \#m(StNN) = &\frac{p}{2}(L-1)Mr + \frac{23}{4}pM(L-1)
    \label{ameq}
\end{align} 
where $M >> p$.
\end{proposition}

\begin{proof}
Using the number of additions and multiplication counts in computing the $[F_i]_N$ by an $N$ dimensional vector and the equations (\ref{eq:recursive_F_1}) and (\ref{eq:recursive_F_2}), the addition and multiplication counts of the StNN can be calculated as follows (assuming \(\log(M)\) recursive factorization steps). 
For each submatrix $i$,
\begin{align*}
    \#a( \hat{D}_i) &= 0, &\#m( \hat{D}_i) &= M \\
    \#a(J) &= 0, &\#m(J) &= 0 \\
    \#a( [F_i]_{2M}) &= 2Mr + 4M, &\#m( [F_i]_{2M}) &= Mr+5M \\
    \#a( \Breve{D}_i) &= 0, &\#m( \Breve{D}_i) &= 2M
\end{align*}

Using the above counts, arithmetic complexity for each submatrices $w^{(1)}_i$  and $w^{(4)}_i$ can be computed.
\begin{align*}
    \#a(w^{(1)}_i) &= 2Mr + 4M, &\#a(w^{(4)}_i) &= 2Mr + 4M\\
    \#m(w^{(1)}_i) &= Mr+8M &\#m(w^{(4)}_i) &= Mr + 6M
\end{align*}
We recall that the $W^{(1)}$ and $W^{(4)}$ contain $p$ number of $w_{i}$ submatrices, introducing $p(2Mr + 4M)$ additions and $p(Mr+8M)$ multiplications for $W^{(1)}$ and $p(2Mr + 4M)$ additions and $p(Mr+6M)$ multiplications for $W^{(4)}$. 
Moreover, arithmetic complexities for $W^{(2)}$ and $W^{(3)}$ can be computed as follows.
\begin{align*}
    \#a(W^{(2)}) &= 2pM,  &\#a(W^{(3)}) &= 2pM \\
    \#m(W^{(2)}) &=4pM,   &\#m(W^{(3)}) &= 2pM
\end{align*}
Next, incorporating bias vectors and computing activation introduces $2pM$ additions and multiplications per each hidden layer, and $M$ additions and $0$ multiplications for the last layer. Additionally, in the last layer, adding the resultant $p$ number of $M$ sized vectors introduces $(p-1)M$ additions. Finally, if one expands the StNN over any number of $L$ layers, where $L-1$ is a multiple of 4 (i.e $L-1 = 4\delta$, $\delta \in \mathbb{Z}^+$), we can repeat the above-described block structure until the last layer. This results in multiplying the total count by $\frac{(L-1)}{4}$. Therefore, the total number of additions and multiplication counts for the StNN is given via (\ref{ameq}).
\end{proof}

\emph{Therefore, with StNN the complexity can be reduced from $\mathcal{O}(M^2L)$ to $\mathcal{O}(pLMlog(M))$, where, $M$ is the number of input and output nodes, $L$ is the number of layers in the neural network and $p$ is the number of submatrices per layer.
}

\begin{remark}
\label{reeq}    
The MSE performance in Section \ref{sec: Simulation} shows that there is a need to adjust and potentially reduce the number of recursive factorization steps into $\lambda (< r) \in \mathbb{N} $ to reduce the MSE values to the order of $10^{-4}$. Although the recursive factorization can be used to reduce the number of learnable weight matrices in the StNN, utilizing this can result in an under-parameterized model, especially when the number of weights becomes insufficient to reduce the MSE. To overcome this challenge, we reduce the number of factorization steps $(\lambda)$ as $N$ increases. Additionally, when we factorize runs up to $\log M$ recursive steps, we may encounter a vanishing gradient problem. This issue arises as the last weights in the factorization step (i.e. $[F_i]_{2}, [F_i]_4$) may not be updated during backpropagation due to very small gradients. However, this can be partially overcome with proper weight initialization techniques\cite{Goodfellow-et-al-2016}. Therefore, reducing factorization steps to $\lambda$ steps allows to improve the overall performance of the model. Hence, computational complexity of the StNN with $\lambda$ recursive steps can be derived from (\ref{ameq}) via
\begin{align}
    \#a_{\lambda}(StNN) = &\frac{p(L-1)M^2}{2^{\lambda-1}} + p\lambda(L-1)M + \frac{3}{4}p(L-1)M
    \label{eq:general FLOP add}
\end{align}
\begin{align}
    \#m_{\lambda}(StNN) = &\frac{(L-1)pM^2}{2^{\lambda-1}} +3Mp(L-1) \nonumber \\ &+ \frac{3}{4}p\lambda(L-1)M,
    \label{eq:general FLOP mul}
\end{align}
We note here that the optimal number of steps $\lambda$ are determined through empirical evaluation and tuning based on the specific characteristics of the MSE requirement and dataset as shown in the numerical simulation followed by the Table \ref{tab:MSE vs N} and \ref{tab:arithmatic_counts} values in the next Section. 
\end{remark}

\section{Simulation Results}\label{sec: Simulation}
In this section, we present numerical simulations based on the StNN to realize wideband multi-beam beamformers. The scaled DVM  $\tilde{A}_N$ by the input vector $\tilde{x} \in \mathbb{C}^N$ result in the output vector $\tilde{y} \in \mathbb{C}^N$ in the Fourier domain. Thus, we show numerical simulations to assess the accuracy and performance of the StNN model in realizing wideband multi-beam beamformers.

\subsection{Numerical Setup for Wideband Multi-beam Beamformers}
{Using an $N$-element uniform linear array (ULA), we could obtain received signals based on the direction of arrival $\theta$, measured counter-clockwise from the broadside direction. The received signals $u_k(t); k=1, 2, ... N$ are defined in the complex exponential form s.t.}
\begin{align}
    u_k(t) = e^{-2j\pi f(t-\Delta t_k)} + n_{s}(t),
\end{align}
where $f$ is the temporal frequency of the signal, 
$t$ is the time at which the signal is received, $\Delta t_k$ denotes the time delay at the $k^{th}$ element of the antenna array,
$n_s(t)$ is complex-valued additive white Gaussian noise (AWGN) with mean $0$ and standard deviation of $0.1$. Moreover, the time delay $\Delta t_k$ is expressed as follows:

\begin{align}
    \Delta t_k = \frac{(k-1) d \sin{\theta}}{c},
\end{align}

where $d=0.5$ represents the antenna spacing, $c$ is the speed of light, 
and $\theta$ stands for the angle of arrival. To train the StNN model, we utilized a dataset that consisted of the sample size of $S := 10000$, 
time-discretized values from $t=0$ to $t=1$ for each antenna array. At time $t_s$, the input vector is determined by the values of $\tilde{x}\in\mathbb{C}^{N}$, and $k = 1, 2, ...,N$ corresponding to the $k^{th}$ element of the antenna array. 
Consequently, the data set can be represented as follows.
\[
X_{S,N} = 
\begin{bmatrix}
  x_1(t_1) & x_2(t_1) & x_3(t_1) & ... & x_N(t_1) \\
  x_1(t_2) & x_2(t_2) & x_3(t_2) & ... & x_N(t_2) \\
  . & . & . & & .\\
 . & . & . & & .  \\
  . & . & . & & .  \\
  x_1(t_S) & x_2(t_S) & x_3(t_S) & ... & x_N(t_S) \\
\end{bmatrix}
\]
The input vector at time $t_s$ for the StNN can be extracted as $\underline{x}(t_s)$ from each row of $X_{S,N}$, where $s$ is a sample size extracted from $S$.

\[
\underline{x}(t_s) = 
\begin{bmatrix}
    x_1(t_s) & x_2(t_s) & x_3(t_s) & ... & x_N(t_s)
\end{bmatrix}
\]

The StNN is then trained with the values of $\underline{x}(t_s)$ to predict the output $\underline{y}(t_s)$. The output vector $\underline{y}(t_s)$ is computed by multiplying $\underline{x}(t_s)$ with the scaled DVM $\widetilde{A}_N$. 
The StNN is trained to predict the result of multiplying the input vector by the DVM.
\[
\underline{y}(t_s) = 
\begin{bmatrix}
    y_1(t_s) & y_2(t_s) & y_3(t_s) & ... & y_N(t_s)
\end{bmatrix}
\]
\begin{align}
    \underline{y}(t_s)^T &= \widetilde{A}_N \times \underline{x}(t_s)^T
\end{align}

Each element in the $\widetilde{A}_N$ can be defined using $\alpha$'s, where $\alpha=e^{-2j\pi f \tau}$. The frequency $f$ is taken as 24 GHz, 27GHz, and 32GHz, and $\tau$ value can approximately be calculated using \cite{ref6}. 
\begin{align*}
    \tau = \frac{2\Delta x}{cN} \approx \frac{1}{f_{max}.N}
\end{align*}

where $\Delta x$ is the antenna spacing and $c$ is the speed of light. In our scenario, \(f_{\text{max}}\):=32 GHz, represents the maximum frequency of the signal. 

\subsection{Numerical Simulations in Realizing Wideband Multi-beam Beamformers}
Here, we discuss the numerical simulations of the StNN to realize wideband multi-beam beamformers. To demonstrate that the StNN model has lower computational complexity compared to FFNN, we conducted numerical simulations of the StNN and FFNN. We standardized the parameters and metrics for both models to ensure a fair comparison. The input layer of both networks comprises $2N$ neurons, representing the real and imaginary parts of elements in the antenna array. Similarly, the output layer has the same number of nodes as the input layer. The compared FFNN includes both a delay layer and a skip connection layer. However, the weight matrices connecting the input layer to the first hidden layer and the last hidden layer to the output layer are both fully connected weight matrices without any structure imposed. We first examine the performance of the StNN for 3 frequencies, i.e., 24GHz, 27GHz, and 32GHz in the range of 24GHz to 32GHz with the receiving signals at 3 different angles (i.e. $\theta = 30, 40 \text{ and } 50$). We generate 1,000 data samples for each angle, resulting in a total of 3,000 data samples for each frequency. Before training the StNN, we split the dataset into 80\% for training and 20\% for validation. For each frequency, we train separate StNN models to evaluate their performance. We conducted simulations for three antenna sizes: \(N = 8\), \(16\), and \(32\). In particular, StNN models with more hidden layers tend to require more epochs and time to converge to an MSE of $10^{-4}$ compared to the small number of hidden layers due to the increased number of weights and model complexity. Additionally, since the relationship between the input features and the target variable is relatively straightforward, the three hidden layer architecture discussed in Section \ref{sec: Methodology} is often sufficient to capture the underlying patterns. Adding more layers introduces unnecessary complexity, leading the model to struggle with generalization. Moreover, training deeper networks requires more computational resources and time \cite{Goodfellow-et-al-2016}. 
Therefore, we adhered to the discussed hidden layer architecture while increasing $p$ in each hidden layer for enhanced convergence. All subsequent simulations for StNN and FFNN use the Leaky-Relu activation\cite{Maas2013RectifierNI} function with $0.2$ scaling factor.
During training, we used the MSE as the loss function and the Levenberg-Marquardt algorithm \cite{more2006levenberg} as an optimization function to learn and update the weights. All the numerical simulations were done in Python (version - 3.10) and Pytorch (version - 2.5) framework to design and train the neural networks. 

\begin{remark}
    To improve readers' comprehension of the theoretical foundation and its relation to the proposed StNN architecture, we encourage readers to access the codes at \href{https://github.com/Hansaka006/Intelligent-Wideband-Beamforming-using-StNN}{\textit{Intelligent Wideband Beamforming using StNN}}.
\end{remark}

\begin{table*}
\renewcommand{\arraystretch}{1.5} 
\caption{This shows MSE values of StNN and FFNN having different antenna array elements. These values are obtained using codes written in {\it Python (Version-3.10)} along with the {\it Pytorch (version-2.5.1)} framework. The term "Model" consists of five numbers representing nodes in input, 3-hidden, and output layers(The first hidden layer is a fully connected layer, the second hidden layer is a delay layer, the third is a skip connection layer, and the last is another fully connected layer.). The notations $p$ and $\lambda$ denote the number of submatrices and recursive steps, respectively. The last column shows the percentage of savings on utilizing StNN over FFNN, leading to a lightweight NN.}
\centering
\begin{tabular}{|c|c|c|c|c|c|c|c|}
\hline
N & Model/ Weights(FFNN)               & MSE (FFNN)                     & Model/ p/ $\lambda$/ Weights(StNN)  & MSE (StNN)                     & $Pr$(Weights)\\ \hline
8   &  $(16, 32, 32, 32, 16) / 1104$      & $(2.8 \pm 0.8) \times 10^{-13}$       & $(16, 32, 32, 32, 16)/ 1/ 4/ 220$         & $(5.6 \pm 0.2) \times 10^{-8}$      & $83\%$  \\ \hline
16  &  $(32, 64, 64, 64, 32) / 4256$        & $(2.8 \pm 4.1) \times 10^{-12}$       & $(32, 64, 64, 64, 32)/ 1/ 5/ 428$       & $(2.0 \pm 0.8) \times 10^{-4}$    &    $90\%$\\ \hline
32  &  $(64, 128, 128, 128, 64) / 16704$     &$(3.2 \pm 0.9)\times 10^{-12}$     & $(64, 128, 128, 128, 64)/ 1/ 6/ 716$        & $(1.0 \pm 3.4)\times 10^{-4}$   &  $96\%$\\ \hline
\end{tabular}
\label{tab:MSE vs N}
\end{table*}
\begin{table}[!t]
\renewcommand{\arraystretch}{1.5}
\caption{Addition and Multiplication counts(FLOPs) for the StNN and FFNN, i.e., FLOPs: = $\#a(StNN) + \#m(StNN)$. The last column shows the percentage of the savings on utilizing StNN (executing $\lambda < r $ recursive steps) over FFNN.}
\centering
\begin{tabular}{|c|c|c|c|}
\hline

N &  \multicolumn{1}{p{1.58cm}|}{\centering FLOPs(FFNN)\\[0.2ex]} & \multicolumn{1}{p{1.55cm}|}{\centering FLOPs(StNN)\\[0.2ex](Eq.(\ref{eq:general FLOP add}) + Eq.(\ref{eq:general FLOP mul}))} & $Pr$(FLOPs)\\ \hline
8        & 2240                       &  992              & 56\%          \\ \hline
16       & 8576                  & 2176                &  75\%             \\ \hline
32       & 33536                   &4736                & 85\%          \\ \hline
\end{tabular}
\label{tab:arithmatic_counts}
\end{table}

\begin{figure*}
\centering
\begin{tabular}{cc}
  \includegraphics[width=86mm]{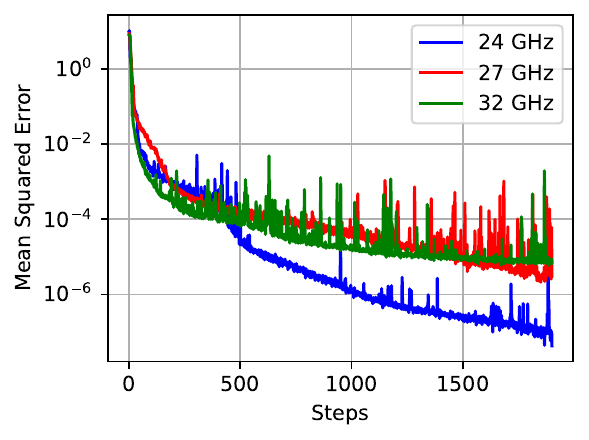} & 
  \includegraphics[width=86mm]{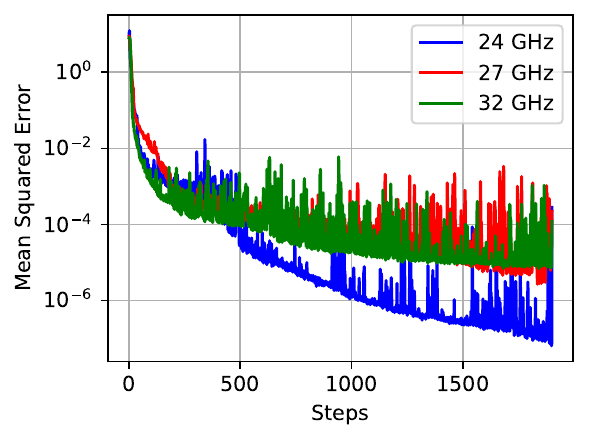}\\
(a) Training performance & (b)  Validation performance
\end{tabular}
\caption{The figures (a) and (b) show training and validation results of the StNNs based on the different frequencies (i.e. 24GHz, 27GHz, and 32GHz) for $N = 16$. These graphs are obtained referencing the "Models" listed in Table \ref{tab:MSE vs N}. When StNN is executed for 1900 epochs, it converges to MSE values of $10^{-5}$. 
These graphs are obtained using {\it Python (Version-3.10)} along with the {\it Pytorch (version-2.5)} framework, and compiled with {\it Levenberg-Marquardt} optimizer.}  
\label{fig:MSEvEpochs_SNN_ANN_Compare}
\end{figure*}

For StNN, we used the scaled DVM factorization in the weight matrix, enabling the training of sparse matrices with a reduced number of weights and FLOPs. The MSE values for the NN predictions in the training and validation sets for the StNN model are shown in Fig. \ref{fig:MSEvEpochs_SNN_ANN_Compare}. We trained both FFNN and StNN models to reach a minimum MSE value between $ 10^{-4}$ and $10^{-3}$. Next, we list and compare the accuracy and performance results of the best models, i.e., low MSE, weights, and FLOP counts, StNN and FFNN models in Table \ref{tab:MSE vs N} and Fig.\ref{fig:MSEvEpochs_SNN_ANN_Compare}. In Table \ref{tab:MSE vs N}, the FFNN and StNN models are conceptualized by the model representing numbers, say-$(A, B_1, B_2, B_3, C)$ s.t. A for the input nodes, B for the hidden nodes, and C for the output nodes. Here, \(p\) and \(\lambda\) denote the number of submatrices and recursive steps, respectively. The final column of Table \ref{tab:MSE vs N} provides information on the percentage reduction (\(Pr\)) of the StNN compared to the FFNN. The \(Pr\) is calculated using the formula \(Pr = \frac{W_{FFNN} - W_{StNN}}{W_{FFNN}} \times 100\%\), where \(W_{FFNN}\) and \(W_{StNN}\) denote the total trainable weights of FFNN and StNN, respectively. However, as shown in Fig. \ref{fig:MSEvEpochs_SNN_ANN_Compare}, when training FFNN and StNN models for 1900 steps(i.e. 380 steps per one epoch and training over 5 epochs), they converge to the MSE values of $10^{-10}$ and $10^{-4}$, respectively. This shows that there is a challenge in maintaining complexity and accuracy simultaneously. Thus, to obtain the MSE with an accuracy of $10^{-4}$, we trained the StNN for 1900 epochs. The main reason is that FFNN models have more weights, which allows for more flexibility during backpropagation, whereas StNN models have fewer weights with the imposed structure. However, The primary advantage of StNN over FFNN is that it has lower arithmetic and space complexity than FFNN making ultimately requires low adders and multipliers for analog and digital intelligent wideband realizations. Furthermore, Table \ref{tab:arithmatic_counts} shows the percentage of FLOP savings for StNN, which can reduce almost 70\% of FLOPs for larger sizes due to the recursive algorithm when compared to FFNN.\\

The simulation results depicted in Table \ref{tab:MSE vs N} highlight a significant trend: as \(N\) increases, the StNN model demonstrates a substantial reduction in weights, leading to a significant decrease in FLOPs as shown in Table \ref{tab:arithmatic_counts}. For larger \(N\) values, such as 32 the StNN model achieves a $95\%$ reduction in weights with an MSE of $10^{-4}$ compared to the FFNN model. As \(N\) increases, it becomes crucial to adjust the value of \(p\) based on the recursive steps \(\lambda\). Increasing more nodes in the hidden layer becomes necessary to reduce MSE and enable the StNN to capture more features \cite{Goodfellow-et-al-2016}.\\

\emph{The findings suggest that incorporating more hidden units can further reduce the MSE. However, it is crucial to note that larger \(p\) values may lead to overfitting, highlighting the importance of selecting the optimal \(p\) value based on the given \(N\).}\\

As depicted in Table. \ref{tab:MSE vs N}, it is evident that for smaller sizes of $N$ (i.e., $N = 8, 16$), performing all the recursive factorization steps is feasible without compromising accuracy. However, with an increase in $N$, conducting multiple recursive factorization steps results in a significant reduction of weights in the network. Unfortunately, this reduction leads to an increase in the MSE, indicating that the StNN model struggles to capture the patterns between input and output. Consequently, for larger $N$ values, it becomes crucial to decrease the recursive steps ($\lambda$) to achieve a lower MSE. In summary, both $\lambda$ and $p$ act as hyperparameters that need to be tuned based on accuracy requirements. 

We note here that our previous work on the S-LSTM network for multi-beam beamformers \cite{ASPSKAL24}, 
saved 30\% of training weights to achieve an MSE of $8\times10^{(-2)}$ for $N = 16$ elements antenna array. Although the S-LSTM approach outperformed conventional LSTM beamforming algorithms, the complexity of the S-LSTM remained relatively high due to the large number of parameters as opposed to the StNN. Thus, in this paper, we showed that the StNN reduces 90\% training parameters compared to FFNN, achieving a significantly lower MSE of $2\times10^{(-4)}$. Such results indicate that our approach better generalizes to intelligent wideband multi-beam beamformers with reduced computational overhead, making it more suitable for real-time hardware-optimized implementations.

Furthermore, the ability to achieve such performance gains with reduced weights and FLoP counts opens pathways for deploying AI-driven wideband multi-beam beamformers in resource-constrained environments. Future work will explore the applicability of this approach to larger antenna array elements, i.e., 128, 256, 256, as well as its adaptability to intelligent signal delaying in nonlinear and time-varying beamforming scenarios.




\section{Conclusion}\label{sec: Conclusion}
We introduced a novel structured neural network (StNN) to intelligently realize wideband multi-beam beamformers utilizing structured weight matrices and submatrices. The proposed StNN leverages the factorization of the DVM in our previous work to reduce the computational complexities of matrix-vector multiplications in the layers of neural networks. Numerical simulation within the range of 24 GHz to 32 GHz shows that the StNN can be utilized to accurately realize wideband multi-beam beamformers as opposed to the conventional fully connected neural network with the complexity reduction from $\mathcal{O}(M^2L)$ to $\mathcal{O}(p L\: M \log M )$, where $M$ is the number of nodes in each layer of the network, $p$ is the number of submatrices per layer, and $M > >  p$. Numerical simulations conducted within the 24 GHz to 32 GHz range have shown that the proposed structured neural architecture can efficiently, accurately, and intelligently be utilized to realize wideband multi-beam beamformers.

\bibliographystyle{IEEEtran}
\bibliography{references}

\end{document}